\documentclass{acmart}
\usepackage{hyperref}
\usepackage{url}
\usepackage{lipsum}
\usepackage{makecell}
\usepackage{soul}
\usepackage{wrapfig}
\usepackage{subcaption}
\usepackage{pifont}
\AtBeginDocument{%
  \providecommand\BibTeX{{%
    \normalfont B\kern-0.5em{\scshape i\kern-0.25em b}\kern-0.8em\TeX}}}

\setcopyright{acmlicensed}
\acmYear{2024}\copyrightyear{2024}
\acmConference[ACM FAccT '24]{ACM Conference on Fairness, Accountability, and Transparency}{June 3--6, 2024}{Rio de Janeiro, Brazil}
\acmBooktitle{ACM Conference on Fairness, Accountability, and Transparency (ACM FAccT '24), June 3--6, 2024, Rio de Janeiro, Brazil}
\acmDOI{10.1145/3630106.3659014}
\acmISBN{979-8-4007-0450-5/24/06}

\usepackage{xcolor}
\definecolor{ForestGreen}{rgb}{0.13, 0.55, 0.13}

\begin{document}

\title{Fairness without Sensitive Attributes via Knowledge Sharing}

\author{Hongliang Ni}
\affiliation{%
  \institution{The University of Queensland}
  \streetaddress{St Lucia}
  \city{Brisbane}
  \state{Queensland}
  \country{Australia}}

\author{Lei Han}
\affiliation{%
  \institution{The University of Queensland}
  \city{Brisbane}
  \state{Queensland}
  \country{Australia}
}

\author{Tong Chen}
\affiliation{%
 \institution{The University of Queensland}
 \city{Brisbane}
 \state{Queensland}
 \country{Australia}}

\author{Shazia Sadiq}
\affiliation{%
  \institution{The University of Queensland}
  \city{Brisbane}
  \state{Queensland}
  \country{Australia}}

\author{Gianluca Demartini}
\affiliation{%
  \institution{The University of Queensland}
  \city{Brisbane}
  \state{Queensland}
  \country{Australia}}

\renewcommand{\shortauthors}{Ni, et al.}

\begin{abstract}
  While model fairness improvement has been explored previously, existing methods invariably rely on adjusting explicit sensitive attribute values in order to improve model fairness in downstream tasks. However, we observe a trend in which sensitive demographic information becomes inaccessible as public concerns around data privacy grow. In this paper, we propose a confidence-based hierarchical classifier structure called ``Reckoner'' for reliable fair model learning under the assumption of missing sensitive attributes. We first present results showing that if the dataset contains biased labels or other hidden biases, classifiers significantly increase the bias gap across different demographic groups in the subset with higher prediction confidence. Inspired by these findings, we devised a dual-model system in which a version of the model initialised with a high-confidence data subset learns from a version of the model initialised with a low-confidence data subset, enabling it to avoid biased predictions. Our experimental results show that Reckoner consistently outperforms state-of-the-art baselines in COMPAS dataset and New Adult dataset, considering both accuracy and fairness metrics.
\end{abstract}



\maketitle

\begin{figure*}[tb]
\begin{subfigure}{0.49\textwidth}
\includegraphics[width=1\linewidth]{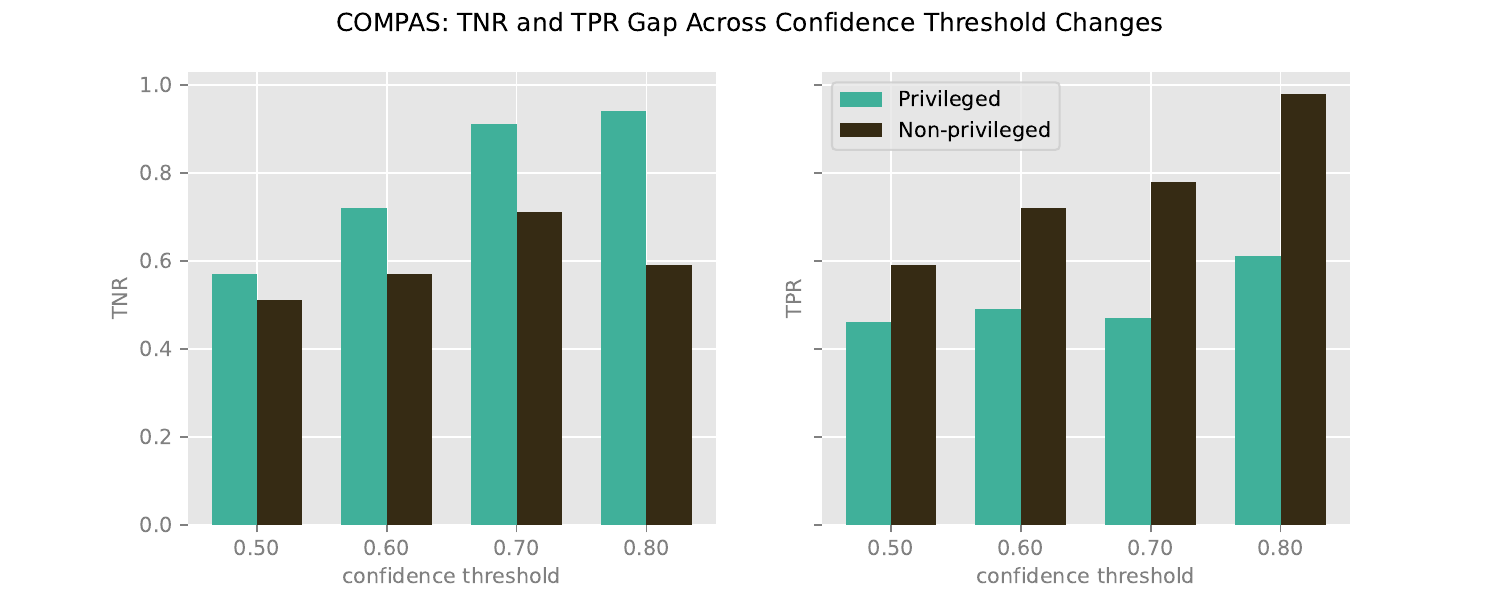}
\caption{}
\label{fig1:a}
\end{subfigure}
\begin{subfigure}{0.49\textwidth}
\includegraphics[width=1\linewidth]{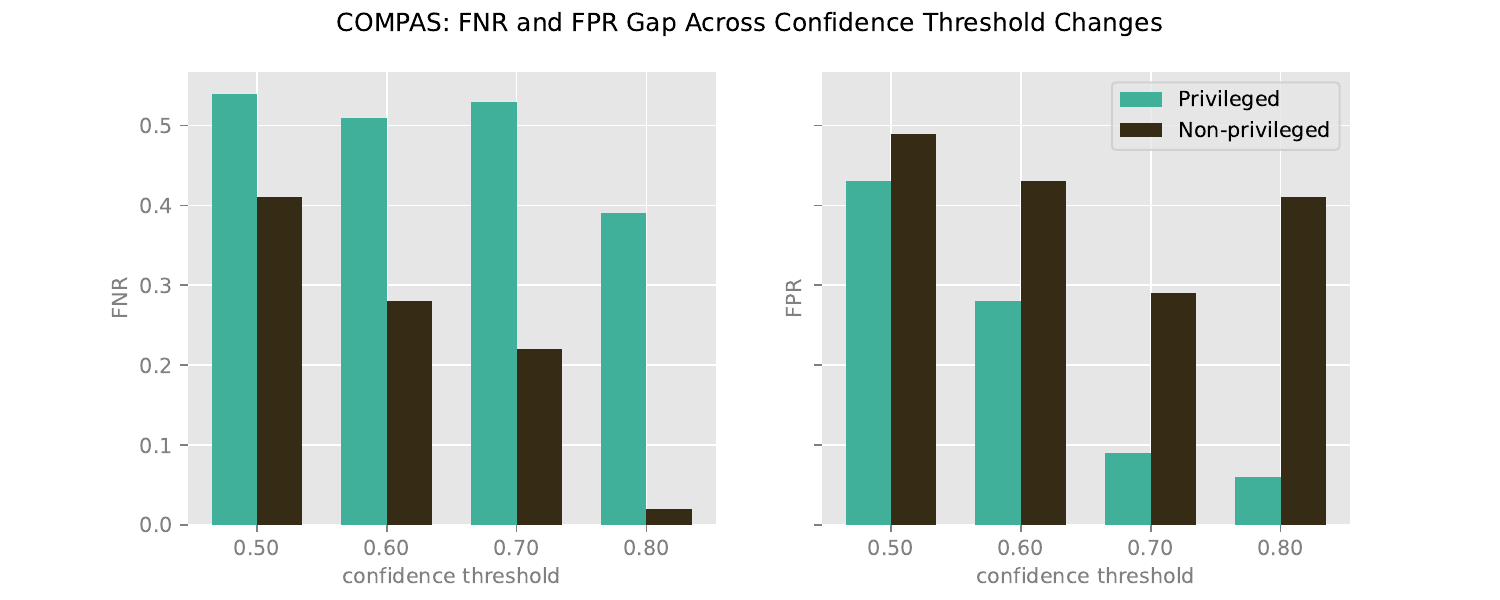}
\caption{}
\label{fig:b}
\end{subfigure}
\begin{subfigure}{0.49\textwidth}
\includegraphics[width=1\linewidth]{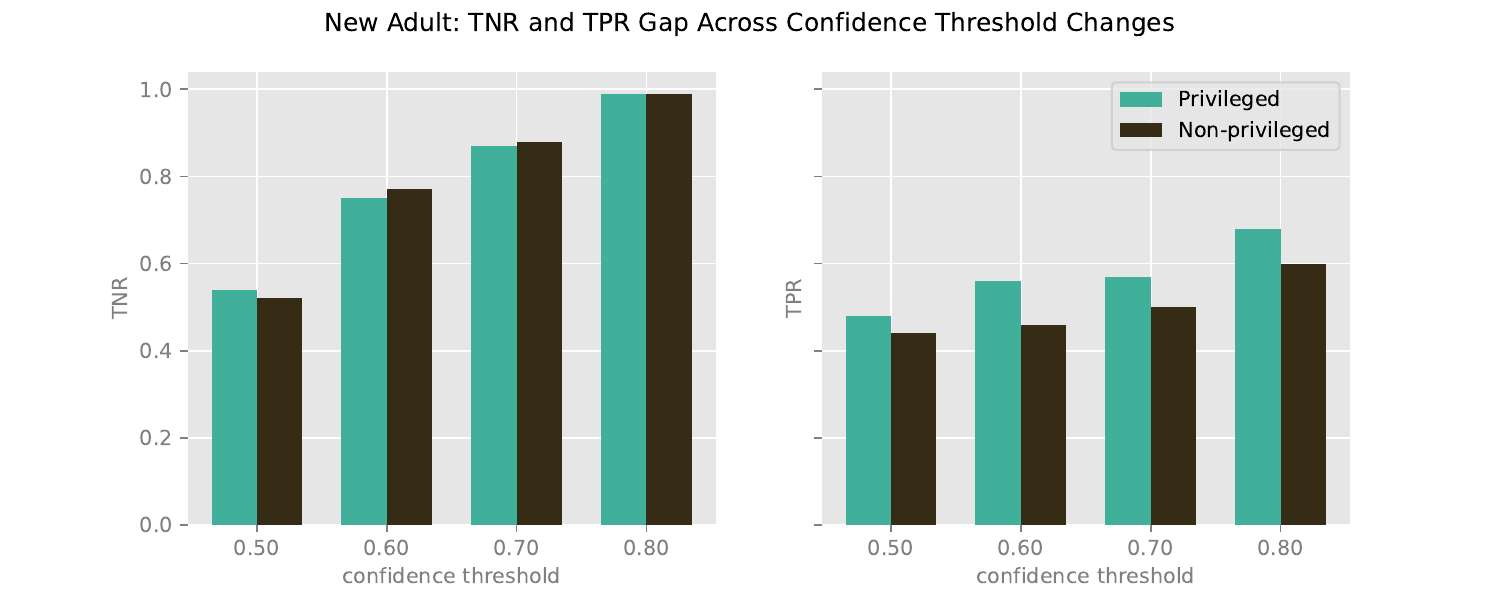}
\caption{}
\label{fig1:c}
\end{subfigure}
\begin{subfigure}{0.49\textwidth}
\includegraphics[width=1\linewidth]{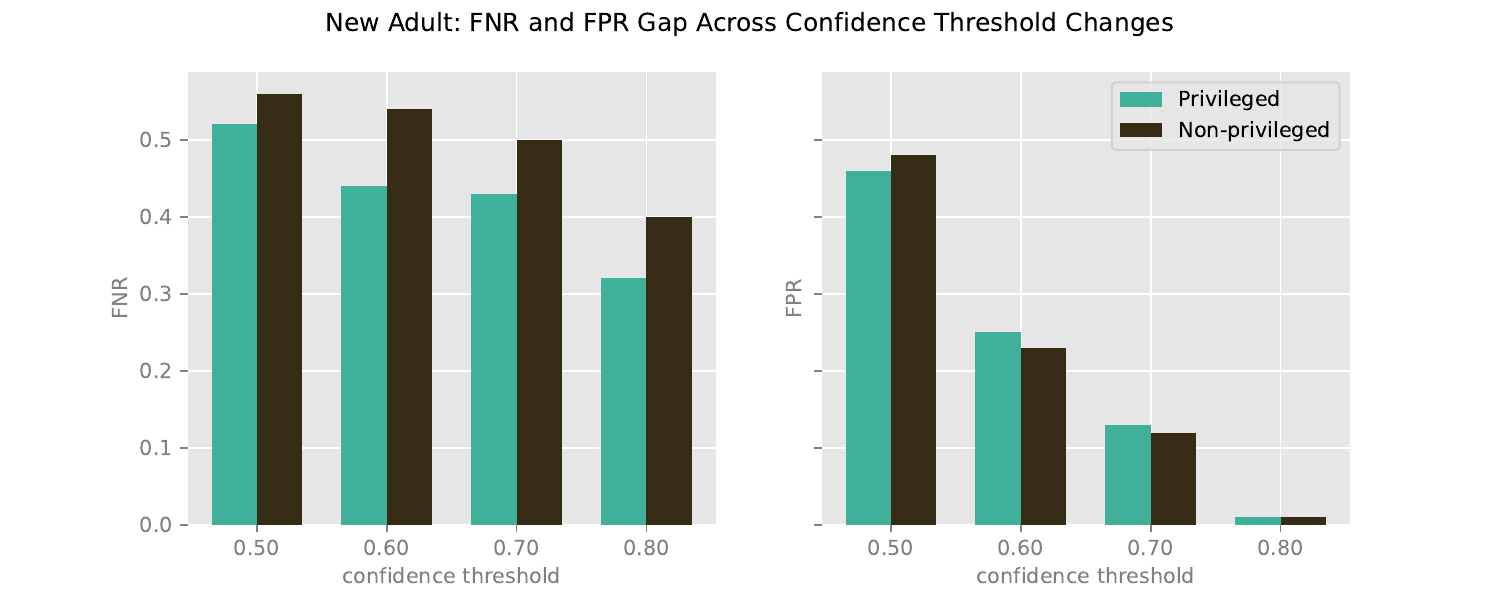}
\caption{}
\label{fig:d}
\end{subfigure}
\caption{Observed measure factor gaps derived from the confusion matrix of the trained logistic regression classifier. (a) and (c) \textbf{True Negative Rate (TNR)} and \textbf{True Positive Rate (TPR)} gaps for two demographic groups across different confidence levels in COMPAS dataset and New Adult dataset, respectively. (b) and (d) \textbf{False Negative Rate (FNR)} and \textbf{False Positive Rate (FPR)} gaps for two demographic groups across different confidence levels in the COMPAS dataset and the New Adult dataset, respectively.}
\Description{Eight histograms, divided into four groups based on two different datasets and four distinct performance indicators of the confusion matrix. Two different colors represent demographic information.}
\label{figure1}
\end{figure*}

\section{Introduction}
Automated models and algorithms have found wide application in various domains, including finance and justice, as tools to assist human decision-making processes \cite{cfpb2022, brennan2009compas}. These applications collect information like age and education level in financial services, or misconduct incidents in policing, as well as sensitive data like race and gender from individuals, raising concerns about the ability of automation to deliver accurate and equitable judgments across diverse demographic groups \cite{andrus2022demographic}. Due to concerns about the misuse of private data, increasing regulatory restrictions have made it more challenging to access and make use of sensitive information in automated decision making \cite{voigt2017eu, bogen2020awareness}. Approaches to improve fairness by not making use of sensitive data can be roughly divided into two categories: (1) focusing on ensuring that accuracy-related utility is equal across various demographic groups \cite{tatsunori2018repeated, preethi200adversarially, wei2023distributionally} and (2) focusing on limiting the impact of sensitivity-correlated proxies on predictions \cite{gupta2018proxy, tianxiang2022related, shen2020class, ghazimatin2022measuring, diana2022multiaccurate}. By relying on the  correlated observed attributes, these methods have the potential benefit of mitigating bias.

However, we argue that fairness is still underachieved because of the unfair data being used.
Most current approaches are limited by the presence of biased labels and other hidden bias concealed within the training data. For example, when faced with judges who are either overly strict or lenient, there may be unusual fluctuations in the number and severity of crimes committed by offenders, leading to biased labels \cite{Eckhouse2017bigdata}. In the supervised learning setting, the model, in an effort to minimize the loss with respect to the biased ground truth, may learn unnecessary biases \cite{preethi200adversarially}. With such data, attempts to obtain auxiliary information, about demographic groups to assist in tasks can unfortunately fail \cite{preethi200adversarially}. Moreover, approaches relying on selected proxy combinations are difficult to  scale for sparse datasets and also difficult to be applied on unstructured data such as image and audio. \cite{datta2017proxy, tianxiang2022related}

In this paper, we aim to answer the following research question: \textit{How can we enhance the fairness of algorithmic predictions when we exclude sensitive information from the dataset?}. We begin by presenting a model confidence study. Confidence calibration is important for classification tasks \cite{guo2017calibration}; however we do not focus on adjusting confidence but rather use the lens of confidence to explore model behaviour. We observe that as the model makes predictions with higher confidence scores, the bias gap between demographic groups increases as well. On the other hand, the analysis of feature value distributions on the COMPAS dataset reveals that, for subsets with higher confidence, the distribution pattern of selected features becomes more distinctive. Combining both analytical results, we argue that in a supervised learning setting, when the model tries to minimise the loss with respect to the training labels, it also learns the bias present in these biased labels. This results in subsets with easier-to-classify samples, yet with predictions which are less fair. 

Inspired by these findings, we initially divide the original training set into two subsets based on confidence scores obtained from a simple linear classifier, and then initialise different classifiers with corresponding data subsets. After that, we introduce learnable noise into the original data, aiming to retain only the necessary information for prediction. In the next phase, one classifier acquires knowledge from the other classifier to learn fairness while also updating itself using the training labels to maintain high levels of effectiveness.

The main contributions of this paper are as follows:
\begin{itemize}
\item By analysing bias and the distribution of non-sensitive attributes across demographic groups in different model confidence intervals, we observe that as the model becomes more certain, it tends to make biased predictions more easily. This analysis reveals how biased labels or other hidden bias adversely affect fairness in machine learning, even when there are no sensitive attributes considered during the supervised learning process. 
\item We introduce a novel confidence-based classification framework, named Reckoner. This framework achieves improved group fairness while maintaining accurate classifications by utilising learnable noise and knowledge-sharing in a dual-model system architecture. This provides an effective approach to improve fairness without using sensitive attributes.
\item We conduct extensive experiments to evaluate the effectiveness of the proposed framework as compared to other baselines in terms of fairness and predictive performance on datasets from which sensitive information is removed. We also present the results of an ablation study to understand the impact on effectiveness of the components in the proposed framework.
\end{itemize}

\begin{figure*}[tb]
\begin{subfigure}{0.49\textwidth}
\includegraphics[width=1\linewidth]{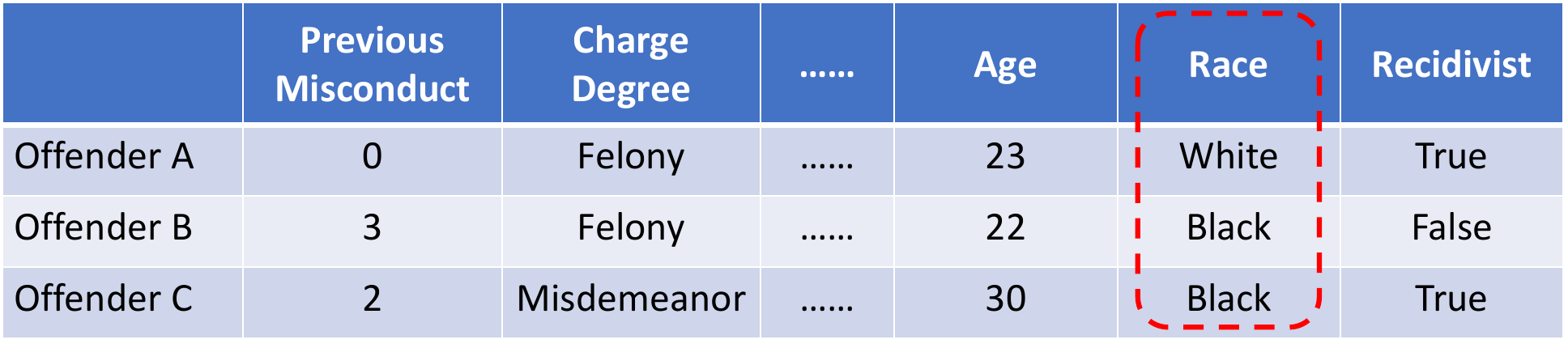}
\caption{}
\label{fig2:a}
\end{subfigure}
\begin{subfigure}{0.49\textwidth}
\includegraphics[width=1\linewidth]{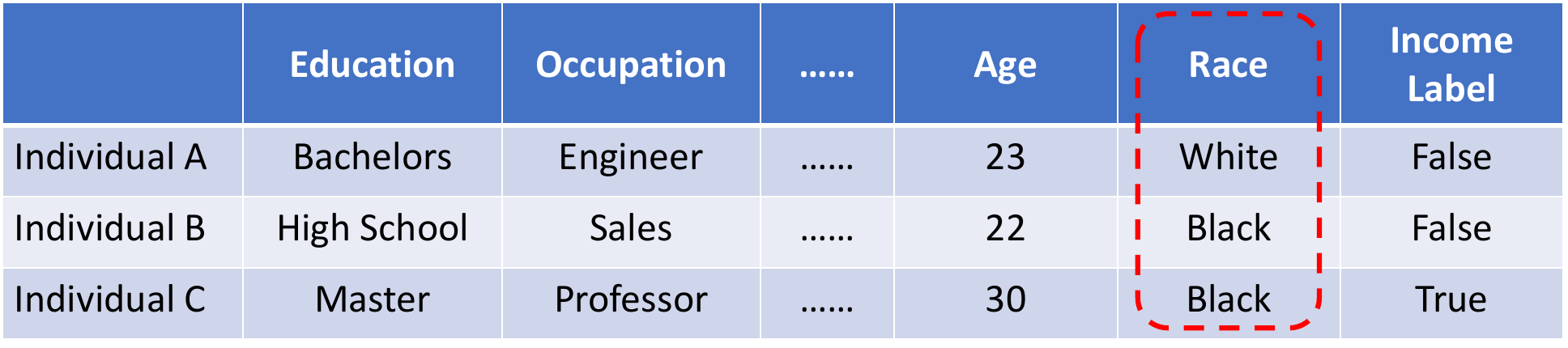}
\caption{}
\label{fig2:b}
\end{subfigure}
\caption{(a) An example of the COMPAS dataset. In our experiment, the attribute ’Race’ in the red box is not used. (b) An example of the New Adult dataset. In our experiment, the attribute ’Race’ in the red box is not used.}
\Description{Two tables represent simple examples of two different datasets.}
\label{figure2}
\end{figure*}

\section{Problem setup and fairness metrics}

\textbf{Problem definition.}
Our goal is to improve group fairness in prediction tasks in a non-sensitive attributes setting, where a set of labeled data $ D = \left \{ x_i, y_i\right \}_{i=1}^{N}$ is available for training. Each $x_i \in \mathbb{R}^{1\times m}$ is a m-dimensional data instance, and we use $F = \left \{f_i, \mathellipsis, f_m\right \}$ to denote the m features. Sensitive attributes S are not used in training, i.e. $S \notin F$. Following the task settings on COMPAS dataset\cite{compas2009}, New Adult dataset\cite{ding2021retiring} and CelebA dataset\cite{liu2015deep}, we focus on binary classification problems, i.e., $y_i \in \{0, 1\}$.
\newline
\textbf{Group Fairness Metrics.} We aim to reduce the difference  in model predictions across various demographic groups. In this paper, we use two group fairness metrics for evaluation: Demographic Parity \cite{corbett2017algorithmic} and Equalised Odds \cite{berk2021fairness}. Both fairness metrics are considered better when they have lower values. 
\begin{itemize}
\item Demographic Parity measures the difference in favourable outcomes between privileged and non-privileged classes:
\begin{equation}
 \Delta_{DP} = p(\hat y = 1 | x_s = s_i) - p(\hat y = 1 | x_s = s_j)
\end{equation}
\item Equalised Odds measures the difference in true positive rates and false positive rates, aiming for equality between privileged and non-privileged classes:
\begin{equation}
\begin{split}
 \Delta_{EOdds} = &\frac{1}{2} |p(\hat y = 1 | x_s = s_i, y = 1) - p(\hat y = 1 | x_s = s_j, y = 1)| + \\
 & \frac{1}{2} |p(\hat y = 1 | x_s = s_i, y = 0) - p(\hat y = 1 | x_s = s_j, y = 0)|,
\end{split}
\end{equation}
 \end{itemize}
where $s_i$ and $s_j$ are arbitrary sensitive attributes from sensitive attribute set $S$ and $\hat y$ is prediction from the classifier. $x_s$ specifies which demographic group data instance $x$ belongs to. In addition to group fairness metrics, we employed the bias gap illustrated in Figure~\ref{figure1} in the confidence analysis in Section~\ref{Section 3}. Taking Figure~\ref{figure1} as an example, we analysed the bias gap of True Positive Rate (TPR) exhibited by two demographic groups: $\Delta_{TPR} = TPR_{s_i} - TPR_{s_j}$. On the other three measure factors, we applied a similar definition to obtain the corresponding bias gap results.

\begin{figure*}[tb]
\centering
\begin{subfigure}{0.49\textwidth}
\includegraphics[width=1\linewidth]{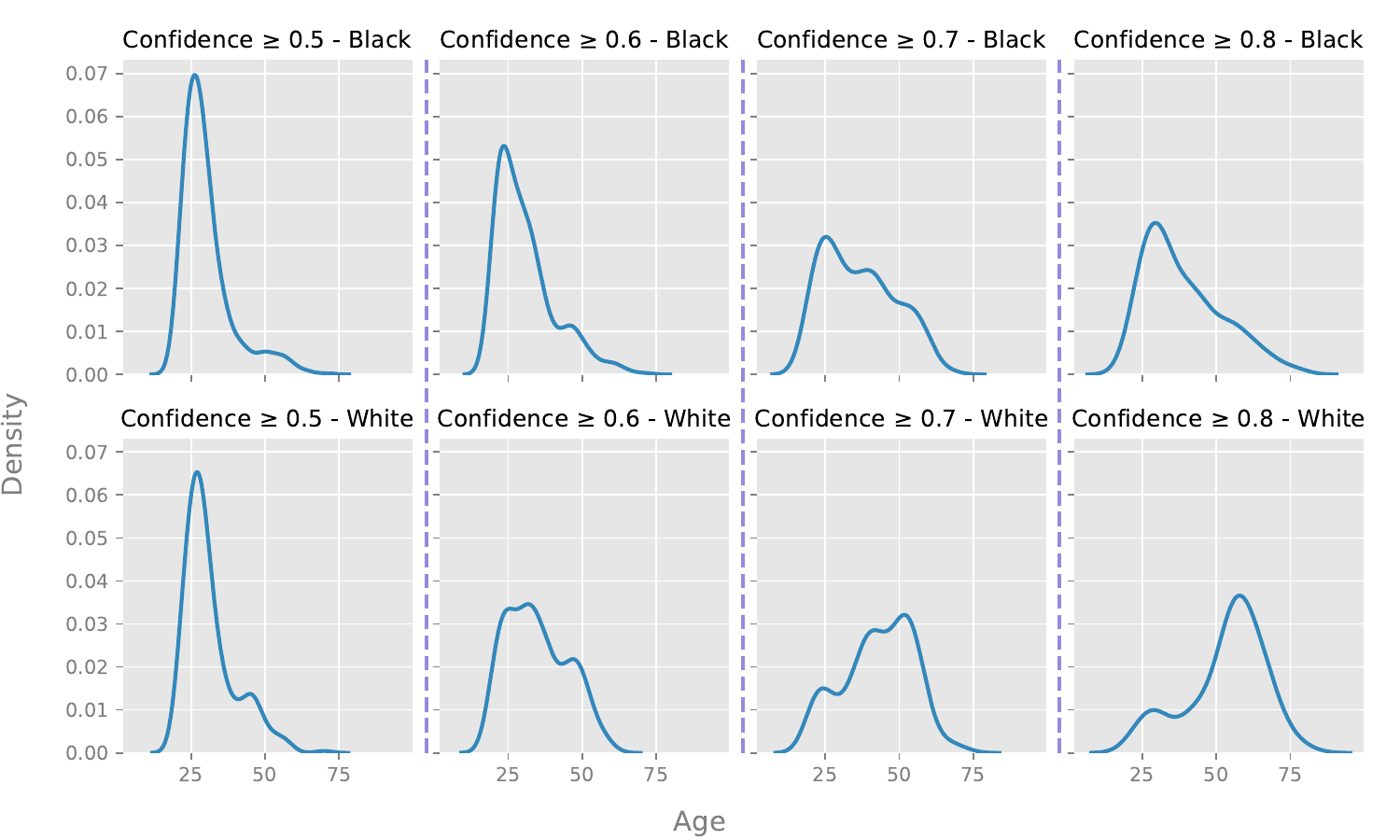}
\caption{}
\label{fig3:a}
\end{subfigure}
\begin{subfigure}{0.49\textwidth}
\includegraphics[width=1\linewidth]{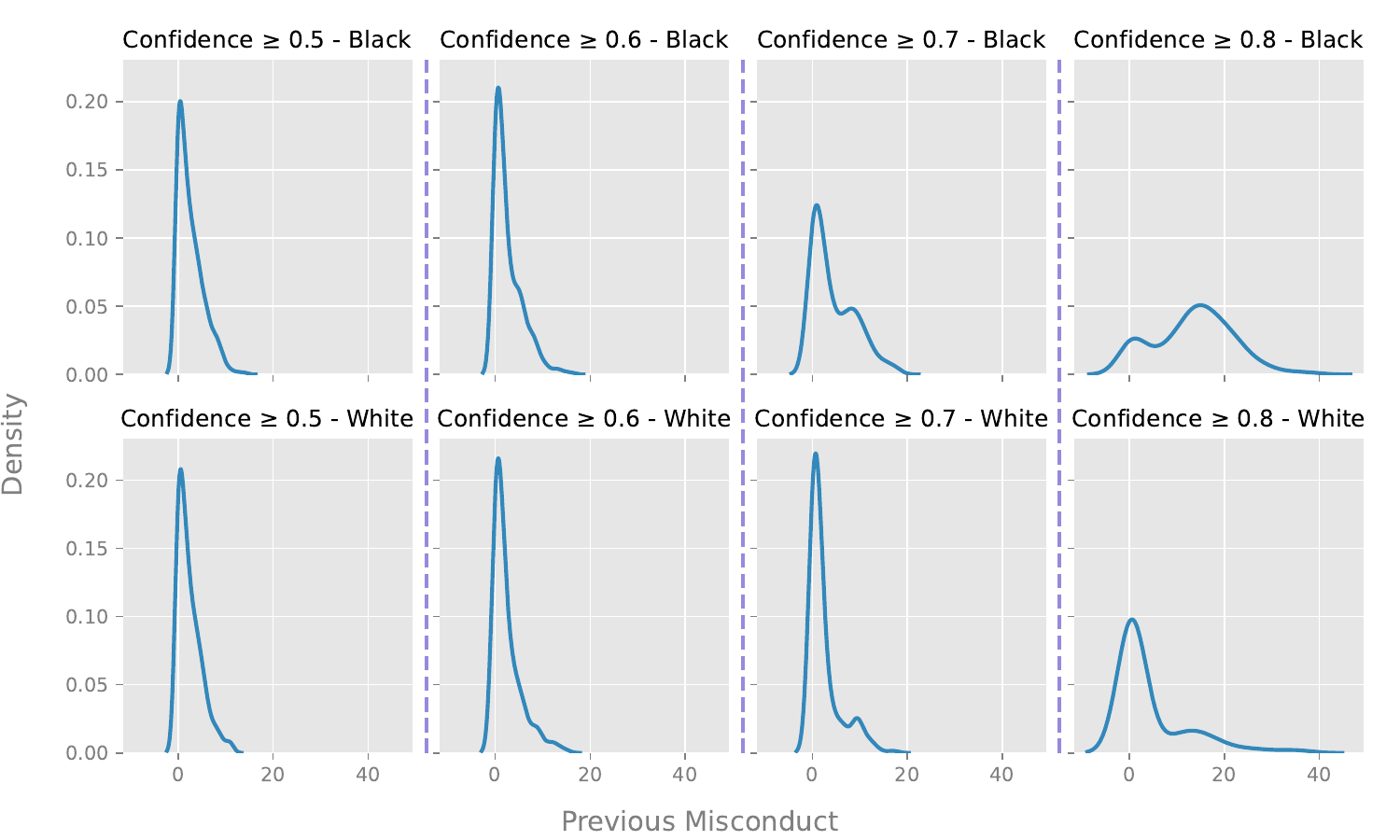}
\caption{}
\label{fig3:b}
\end{subfigure}
\caption{(a) and (b) Distributions of the attribute ’Age’ and 'Previous Misconduct' across different subsets of the testing set.}
\Description{Sixteen distribution plots, divided into two groups based on two different features}
\label{figure3}
\end{figure*}

\section{Analysis of bias gap across confidence levels}
\label{Section 3}

Given current confidence-based approaches, such as Out-of-Distribution (OOD) detection \cite{hendrycks2016baseline} and those for image classification \cite{cui2022confidence, corbiere2019addressing}, we are interested in studying the behaviour of the classifier in subsets with varying levels of confidence. Viewing the data through the lens of confidence provides a more detailed picture of bias patterns across various datasets. In this analysis, we employ logistic regression on two real-world datasets: COMPAS \cite{compas2009} and New Adult \cite{ding2021retiring}. Figure~\ref{fig2:a} provides a toy example using the COMPAS dataset, which is used for predicting offender recidivism. In our setup, sensitive information within the red dashed box is omitted from training. As a consequence, we partitioned the testing dataset into four disjoint subsets using confidence thresholds of 0.5, 0.6, 0.7, and 0.8, respectively. This threshold setting ensures a sufficient number of samples in each subset.

As shown in Figure~\ref{figure1}, for COMPAS dataset, as the confidence threshold increases, the performance differences between demographic groups become more pronounced. Moreover, the gap in FNR and FPR values grows more significantly compared to TNR and TPR values. When the confidence is below 0.6, the gap between the two demographic groups is minimal across all four measures, with the largest bias gap being only 13\%. When the confidence is above 0.8, we expect the classifier in this subset to accurately capture classification patterns. Ideally, there should be high TNR and TPR, the lowest FNR and FPR, and the smallest gap between the two demographic groups. However, the results show that although the first two criteria meet our expectations, the bias gap fluctuates between remarkable values between 35\% and 37\%. Additionally, the classifier has been found to perform better when identifying positive instances in the non-privileged group, as False Negative Rate (FNR) in the privileged group gradually decreases (from 0.54 to 0.39), while it sharply drops from 0.41 to 0.02 for the non-privileged group.

In the New Adult dataset, we can also observe the minimum bias gap in the low-confidence subsets in terms of TPR and FNR. Specifically, both the privileged group and the non-privileged group show an increasing trend in TNR and a sharp decreasing trend in FPR. This indicates that the performance of the classifier in identifying negative samples in both groups  improves, and the performance gaps are small, achieving the ideal performance in the high-confidence subset mentioned above. In contrast, the model struggles in identifying samples where the salary exceeds the set income threshold. Plots of TPR and FNR values also show that bias gaps for these two metrics increase with increasing model confidence. The differences in bias gaps on the two datasets may be attributed to COMPAS being a dataset with biased labels \cite{Eckhouse2017bigdata}. The classifier may mistakenly treat bias as knowledge to learn when it tries to minimise the loss based on the available ground truth labels. Therefore, we observe a significant disparity in bias gap between the low-confidence subset (samples near the decision boundary, confidence $<$ 0.6) and the high-confidence subsets (samples away from the decision boundary, confidence $\geq$ 0.6).

In summary, on both datasets, we observe that bias gaps tends to be relatively smaller in low-confidence subsets for all or some measure factors of classification results, and it increases as confidence levels rise. 
This indicates that models learn to make biased decision with high confidence from the biased labels present in the training data.

To further investigate why the model performs sub-optimally on COMPAS, we selected two non-sensitive attributes from the dataset to understand their distribution patterns across different confidence subsets. Figure ~\ref{fig3:a} and Figure~\ref{fig3:b} show the distribution patterns of the ’Age’ attribute and ’Previous Misconduct’ in various subsets, partitioned based on confidence scores. We can observe that different racial backgrounds exhibit distinguishable distribution patterns within high-confidence subsets, while these differences are not observed in the low-confidence subset. Specifically, within high-confidence subsets, comprising approximately 65\% of the testing data, there are varying tendencies of age distribution dispersion and right-skewness among different racial groups. However,  we only observe a tendency toward right-skewness in age distribution of low-confidence subset. Similar patterns can be observed  in the distribution of previous misconduct. We suspect that this is caused by biased labels and other hidden bias. The classifier, by making biased predictions toward the majority, is enabled to minimise the training loss based on predicting distinguishable patterns from most instances.

\section{Method}

\begin{figure*}[h]
  \centering
  \includegraphics[width=\linewidth]{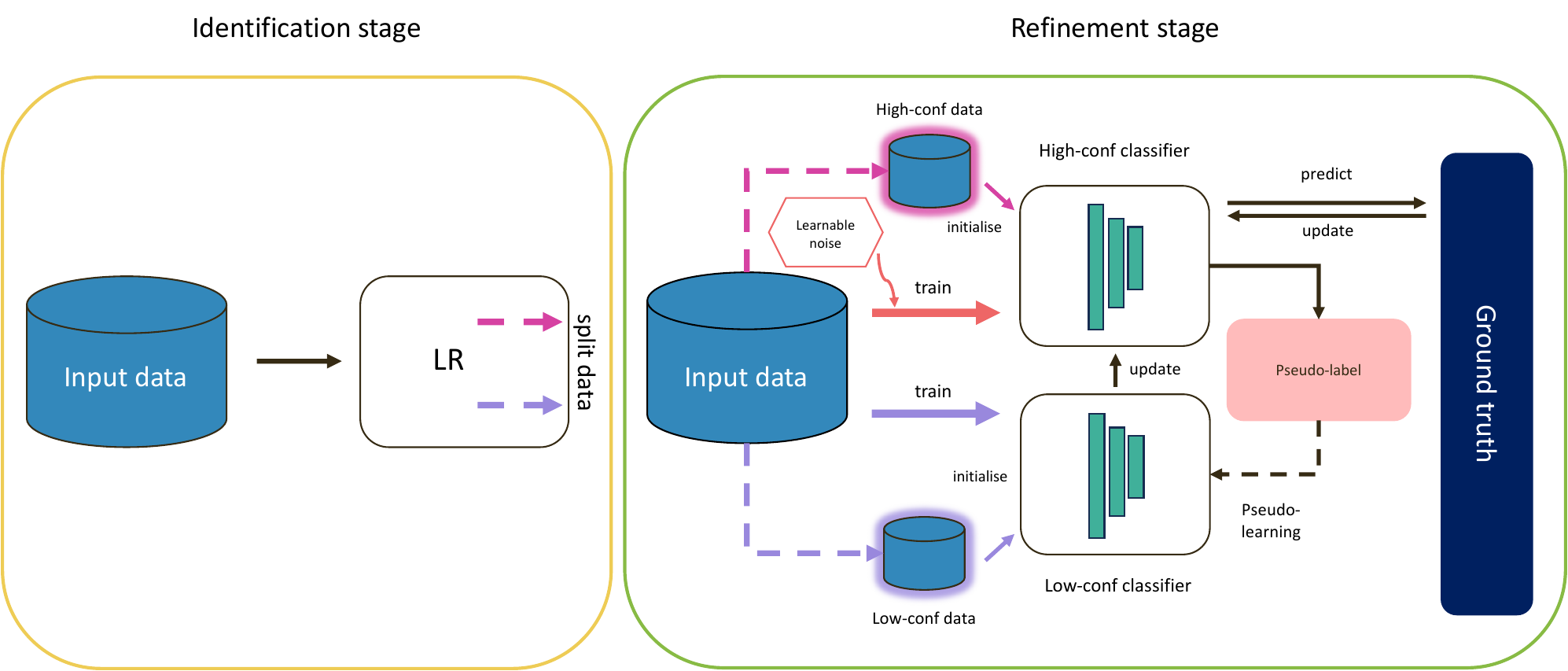}
  \caption{Overview of \textit{\textbf{Reckoner}}. Reckoner consists of two stages. \textit{\textbf{Identification stage}}: we first train a logistic regression classifier on the raw data, and then split the data based on confidence scores. In \textit{\textbf{Refinement stage}}, we introduce learnable noise into the original dataset. We employ two classifiers, one for low-confidence instances and another for high-confidence ones. The \textit{Low-Conf} classifier uses pseudo-labels produced by the \textit{High-Conf} classifier for limited training times and restores for each new data. Knowledge acquired during this process is then shared with the \textit{High-Conf} classifier, which incorporates ground truth data to refine its model weights.}
  \Description{Diagram illustrating the proposed model.}
  \label{figure4}
\end{figure*}

\textbf{Overview.} As shown in Figure~\ref{figure4}, our proposed method consists of two training stages, the Identification stage and Refinement stage. In the Identification stage (Sec.~\ref{Section 4.2}), we employ a simple linear classifier, such as logistic regression, to perform a binary classification task on the raw dataset under a supervised learning setting. Then the training data is split into two subsets based on a predefined confidence threshold: a high-confidence subset and a low-confidence subset. These subsets are then used to initialise their respective classifiers, both of which are three-layer multilayer perceptrons (MLPs). At the beginning of the Refinement stage (Sec.~\ref{Section 4.3}), we introduce learnable noise to the original dataset, generating noise-augmented data for training in this stage. Next, during the iterations, the \textit{High-Confidence} (or ``\textit{High-Conf}'') classifier produces pseudo-labels to train the \textit{Low-Confidence} (or ``\textit{Low-Conf}'') classifier, and the \textit{Low-Conf} classifier updates its knowledge back to the \textit{High-Conf} classifier. The \textit{Low-Conf} classifier is trained for a limited number of iterations, for example, three epochs, before reverting to its initialised state. This approach ensures accurate and unbiased prediction while maintaining an efficient training process. Lastly, the \textit{High-Conf} classifier uses ground truth and shared knowledge to update its parameters.

\subsection{Motivation}
Biased labels and other hidden bias concealed within the dataset results in the classifier learning misleading classification patterns and feature distributions, leading to 1) model-learned parameters that do not enable fair predictions, and 2) limited auxiliary information that non-sensitive attributes can provide. The motivation behind our proposed framework stems from the analytical findings in Section~\ref{Section 3}. We observed the smallest bias gap in the results of the classifier on low-confidence subsets. This insight can be utilised to introduce regularisation to the classifier's learning based on the knowledge acquired from low-confidence subsets. We also introduced learnable noise, to obtain more reliable auxiliary information. To improve fairness in the settings of missing sensitive attributes, we design a novel framework, named Reckoner, which seamlessly integrates learnable noise and a knowledge-sharing mechanism between dual models. We have demonstrated the necessity of combining these two components in our ablation study (see Section ~\ref{sec:ablation}).

\subsection{Identification stage} \label{Section 4.2}
In this stage, we perform a simple confidence-based sample split on the training data to obtain high-confidence samples and low-confidence samples. Specifically, we train a logistic regression classifier on the original training set and we split the data using confidence threshold (in this case, it is set to 0.6 following the results in \cite{himabindu2017uu}). Our hypothesis is that the model trained on the low-confidence subset is more inclined towards classifications with smaller bias gap, even if its predictive accuracy is relatively modest. However, by integrating the knowledge derived from the model trained on the high-confidence subset, it is possible to enhance prediction effectiveness while maintaining good fairness. 

\subsection{Refinement stage} \label{Section 4.3}
\subsubsection{Learnable noise}
In the initial phase of this stage, learnable noise is introduced into the training set. Intuitively, learnable noise has a similar effect to L1 regularisation. L1 regularisation is employed to encourage sparsity in the model parameters, resulting in a model that utilises only a subset of the initial features. This is beneficial for reducing bias gaps in model predictions because, even if the dataset does not contain sensitive information, the associated non-sensitive information may still be  biased. Therefore, employing feature selection methods becomes necessary. However, we do not directly use L1 regularisation here because it is constrained by the model structure and feature scale thus being less generalisable. L1 regularisation imposes a penalty on the weight of each feature, which is the sum of the absolute values of the weights. Consequently, changes to the model structure or adjustments to the feature scale can change the weights, affecting the strength of L1 regularisation. On the other hand, adding learnable noise to input data can be considered as disrupting the original inputs to increase the sparsity. Through supervised learning, the model is encouraged to select features that have  higher cross-entropy with the ground truth. Moreover, learnable noise is robust to changes in model structure or feature scale, allowing us to leverage this property to design a more flexible framework where the classifier can be any model suitable for downstream tasks. To be more specific, we add a noise wrapper to vectors of the same dimensions as the input (denoted as $\eta$). The noise wrapper is a simple two-layer MLP that is subsequently applied to modify the input. The resulting modified input is referred to as a noise-augmented input and can be represented as follows:
\begin{equation}
 \tilde x_{i} = x_{i} + tanh(g_{\omega}(\eta)),
\end{equation}
where $\omega$ is the set of parameters in the noise wrapper $g$, and $tanh(\cdot)$ helps constrain the range of noise values to [-1, 1]. In the rest of the refinement stage, $\tilde x_{i}$ is the new input we use for the \textit{High-Conf} classifier.

\subsubsection{Dual-model and knowledge sharing}
The lower predictive performance prevents us from relying on the \textit{Low-Conf} classifier to perform classification tasks, but it can guide the \textit{High-Conf} classifier to make fairer classifications. Within the Reckoner framework, the \textit{Low-Conf} classifier relies on the pseudo-labels produced by the \textit{High-Conf} classifier for supervised learning to update its own parameters. Since ground truth data is not used in this phase, the learning process can  be referred to as \textit{pseudo-learning}. The reason we do not use ground truth here is that some datasets with hidden bias contain a considerable amount of biased labels which may be propagated during model learning. If these labels were used during the learning phase of the \textit{Low-Conf} classifier, it might lead the model to make biased predictions, thereby limiting its regularisation effect on the \textit{High-Conf} classifier. We use binary cross entropy as the supervised loss of \textit{pseudo-learning}:
\begin{equation}
\mathcal{L}_{\text{L}} = BCE(f_{\Theta_{\text{L}}}(x), \tilde{y}),
\end{equation}

where $f_{\Theta_{\text{L}}}$ is \textit{Low-Conf} classifier and $\tilde{y}$ is the pseudo-label produced by \textit{High-Conf} classifier. The subscripts ``L'' refers to the \textit{Low-Conf} classifier. Note that during this phase, the training iterations of the \textit{Low-Conf} classifier are limited (set to only 3 times in our experiments) for training efficiency. Furthermore, once these iterations end, the \textit{Low-Conf} classifier has a rollback operation, reverting its parameters to their initialised values. This design maintains the effectiveness of \textit{Low-Conf} classifier in providing fairness guidance to the parameters of \textit{High-Conf} classifier, and avoids learning the bias inherent in the dataset.

On the other hand, we rely on the \textit{High-Conf} classifier, which offers higher accuracy, to perform classification tasks. However, as revealed by the analysis in Section~\ref{Section 3}, we are aware of its poor performance in terms of decreasing bias. Previous work shows that leveraging the strengths of both models is a common strategy for various research problems, such as, for example, the high-pass and low-pass filters in graph neural networks \cite{bo2021beyond}, and efforts to average model weights for improved image classification \cite{tarvainen2017mean}. The most promising improvement on fairness for the \textit{High-Conf} classifier lies in integrating the knowledge from the \textit{Low-Conf} classifier, and its parameter update mechanism can be expressed as follows:
\begin{equation}
\Theta_{\text{H}} \leftarrow \alpha\Theta_{\text{H}} + (1-\alpha)\Theta_{\text{L}},
\end{equation}
where $\alpha$ controls the proportion of the knowledge of \textit{High-Conf} classifier. The subscripts ``H'' and ``L'' refer to the \textit{High-Conf} classifier and the \textit{Low-Conf} classifier, respectively. In order to enhance predictive accuracy, we use ground truths and employ the backpropagation algorithm to update the \textit{High-Conf} classifier. By integrating the knowledge from the \textit{Low-Conf} classifier, the final update mechanism can be formulated as follows:
\begin{equation}
\theta_{i}^{\text{H}} \leftarrow \hat{\theta}_{i-1}^{\text{H}} - \gamma\frac{\partial \mathcal{L}_{\text{H}}} {\partial  \hat{\theta}_{i-1}^{\text{H}}}, \quad  \hat{\theta}_{i-1}^{\text{H}} \leftarrow \alpha\theta_{i-1}^{\text{H}} + (1-\alpha)\theta_{k}^{\text{L}},
\end{equation}
where $\theta_{i}^{\text{H}}$ is the weight of \textit{High-Conf} classifier at $i$-th iteration, $\hat{\theta}_{i-1}^{\text{H}}$ is the temporary weight integrating both \textit{High-Conf} classifier and \textit{Low-Conf} classifier knowledge controlled by $\alpha$, and $k$ is the iteration number when the \textit{Low-Conf} classifier achieves the best performance during \textit{pseudo-learning}. We also employ binary cross entropy as the supervised loss of classification task:

\begin{equation}
\mathcal{L}_{\text{H}} = BCE(f_{\Theta_{\text{H}}}(x), y).
\end{equation}

Intuitively, the \textit{pseudo-learning} applied to the \textit{High-Conf} classifier can be interpreted as shifting the  decision boundary closer to the feature space of the the major samples in high-confidence subsets, with the hyperparameter $\alpha$ controlling stability. Hence, the model will not misclassify similar instances based on distribution patterns of the majority. Another component of the framework, learnable noise, offers more auxiliary information for demographic groups, ensuring both accuracy and enhanced fairness. We will discuss the contribution of each of these two components to prediction fairness when discussing the results of the ablation study (see Section \ref{sec:ablation}).

\section{Experimental Evaluation}
\textbf{Datasets.} We validate our model on three benchmark datasets:(1) \textbf{New Adult}: as introduced by \cite{ding2021retiring}, it comprises of 49,531 samples, each associated with 14 attributes. The primary objective is to predict whether an individual’s income exceeds 50k. In our experiments, we convert income into binary labels, and we set `race' as the sensitive attribute and exclude it from our experiments. (2) \textbf{COMPAS}: COMPAS~\cite{compas2009} comprises 7,215 data samples, each associated with 11 attributes. Following previous work on fairness without sensitive attributes~\cite{junyi2022knowledge}, we have filtered this dataset to include only African American and Caucasian offenders, hence we use a modified dataset containing 6,150 samples. The primary objective is to predict whether an offender will commit another offense within two years. We set `race' as the sensitive attribute and exclude it from our experiments. (3) \textbf{CelebA}: The CelebA dataset \cite{liu2015deep} comprises 202,599 image samples with resolution 178*218, each associated with 40 attributes. The primary objective is to predict the attractiveness of each image. We set `gender' as the sensitive attribute and exclude it from our experiments.

\textbf{Baselines.} We compare our method with four related methods for  comparisons: (1) Distributed Robust Optimisation (DRO) \cite{tatsunori2018repeated}: The primary objective of this method is to enhance Rawlsian Max-Min Fairness \cite{rawls2001justice}. It specifically focuses on mitigating the prioritisation of benefits for the majority group that may arise from employing empirical risk minimisation. The approach establishes upper and lower bounds for the objective function of each group based on different group proportions, ensuring non-discrimination by the algorithm. (2) ARL \cite{preethi200adversarially}: This approach also aims at optimising Rawlsian Max-Min Fairness and uses adversary learning to optimise worst-case performance by prioritising instances with higher loss. (3) FairRF \cite{tianxiang2022related}: This approach identifies proxy features strongly correlated with sensitive attributes and minimises the correlation by re-weighting to achieve fairness. (4) \cite{junyi2022knowledge}: This approach applies knowledge distillation requiring one model to produce soft labels, and uses them to train a second model to obtain a better decision boundary. It has two variants: either with labels determined by softmax function or linear function.

\textbf{Experimental Setting.} We employ feature hashing on categorical features if the dataset we  use  contains categorical features. For the proposed framework, we apply logistic regression to train a simple binary classifier and follow \cite{himabindu2017uu} by setting 0.6 as the confidence threshold for data splitting in the identification stage. In the refinement stage, we use each confidence-based subset and use 10\% of the total model training iterations to initialise both the \textit{High-Conf} classifier and the \textit{Low-Conf} classifier. In the \textit{pseudo-learning} phase, the \textit{Low-Conf} classifier is trained three times. In the training for the whole proposed framework, we use Adam to be the optimiser and binary cross entropy for classification loss. We employ Resnet-50 \cite{he2016deep} as the backbone of the classifiers for the CelebA dataset. For evaluation, we use Equalised Odds \cite{berk2021fairness} and Demographic Parity \cite{corbett2017algorithmic} as fairness metrics and report accuracy for classification. Both fairness metrics are considered better when they have lower values.

\begin{table*}[t]
\caption{Results on the COMPAS dataset.}
\Description{Table showing three performance metrics of the proposed method on the COMPAS dataset.}
\begin{center}
\begin{tabular}{ |p{5cm}|p{5cm}|p{5cm}|p{5cm}|}
 \toprule
 \multicolumn{1}{c}{Metrics(\%)} & \multicolumn{1}{c}{Accuracy} & \multicolumn{1}{c}{Equalised Odds} & \multicolumn{1}{c}{Demographic Parity} \\
 \multicolumn{1}{c}{Methods} \\
 \hline
 \multicolumn{1}{l||}{DRO \cite{tatsunori2018repeated}} & \multicolumn{1}{c}{$64.88\pm 0.34\%$} & \multicolumn{1}{c}{$23.11 \pm 1.80\%$} & \multicolumn{1}{c}{$25.32\pm 1.22\%$} \\  
 \multicolumn{1}{l||}{ARL \cite{preethi200adversarially}} & \multicolumn{1}{c}{$65.32\pm 0.70\%$} & \multicolumn{1}{c}{$23.01\pm 1.21\%$} & \multicolumn{1}{c}{$25.37\pm 1.01\%$} \\ 
 \multicolumn{1}{l||}{FairRF \cite{tianxiang2022related}} & \multicolumn{1}{c}{$63.26 \pm 0.83\%$} & \multicolumn{1}{c}{$25.67 \pm 2.63\%$} & \multicolumn{1}{c}{$21.47 \pm 1.76\%$} \\ 
 \multicolumn{1}{l||}{Chai's work \cite{junyi2022knowledge}(softmax label)} & \multicolumn{1}{c}{$63.47 \pm 0.44\%$} & \multicolumn{1}{c}{$21.32 \pm 1.97\%$} & \multicolumn{1}{c}{$19.52 \pm 2.46\%$} \\ 
  \multicolumn{1}{l||}{Chai's work \cite{junyi2022knowledge}(linear label)} & \multicolumn{1}{c}{$63.34 \pm 0.46\%$} & \multicolumn{1}{c}{$20.31 \pm 2.62\%$} & \multicolumn{1}{c}{$20.27 \pm 2.34\%$} \\ 
 \hline
 \multicolumn{1}{l||}{Reckoner} & \multicolumn{1}{c}{$64.92 \pm 0.63\%$} & \multicolumn{1}{c}{$17.47 \pm 0.87\%$} & \multicolumn{1}{c}{$20.72 \pm 0.97\%$}\\

 \multicolumn{1}{l||}{Reckoner (w/o noise)} & \multicolumn{1}{c}{$64.95 \pm 0.51\%$} & \multicolumn{1}{c}{$17.91 \pm 1.32\%$} & \multicolumn{1}{c}{$21.21 \pm 1.33\%$} \\

 \multicolumn{1}{l||}{Reckoner (w/o \textit{pseudo-learning})} & \multicolumn{1}{c}{$64.38 \pm 0.83\%$} & \multicolumn{1}{c}{$17.98 \pm 1.34\%$} & \multicolumn{1}{c}{$21.18 \pm 1.46\%$} \\
 \bottomrule
\end{tabular}
\end{center}
\label{table1}
\end{table*}

\begin{table*}[t]
\caption{Results on the New Adult dataset.}
\Description{Table showing three performance metrics of the proposed method on the New Adult dataset.}
\begin{center}
\begin{tabular}{ |p{6cm}|p{4cm}|p{4cm}|p{4cm}|}
 \toprule
 \multicolumn{1}{c}{Metrics(\%)} & \multicolumn{1}{c}{Accuracy} & \multicolumn{1}{c}{Equalised Odds} & \multicolumn{1}{c}{Demographic Parity} \\
 \multicolumn{1}{c}{Methods} \\
 \hline
 \multicolumn{1}{l||}{DRO \cite{tatsunori2018repeated}} & \multicolumn{1}{c}{$85.15 \pm 0.93\%$} & \multicolumn{1}{c}{$11.56 \pm 2.10\%$} & \multicolumn{1}{c}{$12.23 \pm 1.41\%$} \\  
 \multicolumn{1}{l||}{ARL \cite{preethi200adversarially}} & \multicolumn{1}{c}{$85.37 \pm 1.91\%$} & \multicolumn{1}{c}{$11.79 \pm 1.77\%$} & \multicolumn{1}{c}{$13.05 \pm 1.57\%$} \\
 \multicolumn{1}{l||}{FairRF \cite{tianxiang2022related}} & \multicolumn{1}{c}{$83.74 \pm 0.86\%$} & \multicolumn{1}{c}{$11.23 \pm 1.42\%$} & \multicolumn{1}{c}{$11.37 \pm 1.46\%$} \\ 
 \multicolumn{1}{l||}{Chai's work \cite{junyi2022knowledge}(softmax label)} & \multicolumn{1}{c}{$84.63 \pm 0.47\%$} & \multicolumn{1}{c}{$10.34 \pm 1.22\%$} & \multicolumn{1}{c}{$10.63 \pm 1.34\%$} \\ 
  \multicolumn{1}{l||}{Chai's work \cite{junyi2022knowledge}(linear label)} & \multicolumn{1}{c}{$84.27 \pm 0.31\%$} & \multicolumn{1}{c}{$10.57 \pm 1.64\%$} & \multicolumn{1}{c}{$10.21 \pm 1.52\%$} \\ 
 \hline
 \multicolumn{1}{l||}{Reckoner} & \multicolumn{1}{c}{$85.35 \pm 0.07\%$} & \multicolumn{1}{c}{$5.83 \pm 0.51\%$} & \multicolumn{1}{c}{$9.78 \pm 0.17\%$} \\
 
 \multicolumn{1}{l||}{Reckoner (w/o noise)} & \multicolumn{1}{c}{$85.36 \pm 0.09\%$} & \multicolumn{1}{c}{$5.82 \pm 0.27\%$} & \multicolumn{1}{c}{$9.98 \pm 0.19\%$} \\

 \multicolumn{1}{l||}{Reckoner (w/o \textit{pseudo-learning})} & \multicolumn{1}{c}{$85.53 \pm 0.13\%$} & \multicolumn{1}{c}{$4.82 \pm 0.41\%$} & \multicolumn{1}{c}{$9.11 \pm 0.18\%$} \\
 
 \bottomrule
\end{tabular}
\end{center}
\label{table2}
\end{table*}

\begin{table*}[t]
\caption{Results on the CelebA dataset.}
\Description{Table showing three performance metrics of the proposed method on the CelebA dataset.}
\begin{center}
\begin{tabular}{ |p{6cm}|p{4cm}|p{4cm}|p{4cm}|}
 \toprule
 \multicolumn{1}{c}{Metrics(\%)} & \multicolumn{1}{c}{Accuracy} & \multicolumn{1}{c}{Equalised Odds} & \multicolumn{1}{c}{Demographic Parity} \\
 \multicolumn{1}{c}{Methods} \\
 \hline
 \multicolumn{1}{l||}{DRO \cite{tatsunori2018repeated}} & \multicolumn{1}{c}{$77.12 \pm 0.58\%$} & \multicolumn{1}{c}{$17.22 \pm 1.69\%$} & \multicolumn{1}{c}{$19.04 \pm 1.51\%$} \\  
 \multicolumn{1}{l||}{ARL \cite{preethi200adversarially}} & \multicolumn{1}{c}{$78.91 \pm 0.41\%$} & \multicolumn{1}{c}{$17.53 \pm 1.72\%$} & \multicolumn{1}{c}{$19.46 \pm 1.96\%$} \\
 \multicolumn{1}{l||}{Chai's work \cite{junyi2022knowledge}(softmax label)} & \multicolumn{1}{c}{$80.87 \pm 0.14\%$} & \multicolumn{1}{c}{$11.43 \pm 1.25\%$} & \multicolumn{1}{c}{$15.27 \pm 1.71\%$} \\ 
  \multicolumn{1}{l||}{Chai's work \cite{junyi2022knowledge}(linear label)} & \multicolumn{1}{c}{$80.76 \pm 0.73\%$} & \multicolumn{1}{c}{$10.62 \pm 1.10\%$} & \multicolumn{1}{c}{$14.47 \pm 1.64\%$} \\ 
 \hline
 \multicolumn{1}{l||}{Reckoner} & \multicolumn{1}{c}{$79.47 \pm 0.25\%$} & \multicolumn{1}{c}{$11.58 \pm 0.49\%$} & \multicolumn{1}{c}{$13.92 \pm 1.27\%$} \\
 
 \multicolumn{1}{l||}{Reckoner (w/o noise)} & \multicolumn{1}{c}{$79.86 \pm 0.09\%$} & \multicolumn{1}{c}{$11.96 \pm 0.67\%$} & \multicolumn{1}{c}{$14.23 \pm 0.85\%$} \\

 \multicolumn{1}{l||}{Reckoner (w/o \textit{pseudo-learning})} & \multicolumn{1}{c}{$77.99 \pm 0.11\%$} & \multicolumn{1}{c}{$11.01 \pm 0.74\%$} & \multicolumn{1}{c}{$13.19 \pm 0.94\%$} \\
 
 \bottomrule
\end{tabular}
\end{center}
\label{table3}
\end{table*}

\subsection{Results}
Tables~\ref{table1} - ~\ref{table3} show the results comparing our models to other baselines. Results for both variants of \cite{junyi2022knowledge} and of FairRF \cite{tianxiang2022related} are from \cite{junyi2022knowledge} using the same datasets and same train-valid-test split. Our model outperforms the selected baselines. In the COMPAS dataset, we can observe that Reckoner achieves the best result in Equalised Odds with a relative improvement of about 2.84\% over the best baseline, and also secures the second-best effectiveness in terms of accuracy. Compared to \cite{junyi2022knowledge} with the optimal Demographic Parity, although  our method exhibits a marginal difference of 1.2\%, we obtain improvements in terms of accuracy and Equalised Odds, with improvements of 1.45\% and 3.85\%, respectively. In comparison to the highest accuracy achieved by ARL~\cite{preethi200adversarially}, Reckoner exhibits only a 0.4\% gap in accuracy. However, it holds a significant edge in fairness, with improvements of 5.54\% for Equalised Odds and 4.65\% for Demographic Parity. In the New Adult dataset, Reckoner exhibits  improvements in fairness  compared to all the baselines. In comparison to the best-performing baselines in terms of fairness, it achieves a 4.51\% improvement in Equalised Odds and a 0.43\% improvement in Demographic Parity.
At the same time, Reckoner achieves the second-highest position in terms of prediction accuracy, with a marginal 0.02\% gap compared to the most accurate baseline. In the CelebA dataset \cite{liu2015deep}, on the other hand, we obtain a third position in accuracy and Equalised Odds, but  achieve the best Demographic Parity.

Note that we differentiate our study from those that emphasise equal accuracy-related utility across different demographic groups (such as the percentage of matched binary labels and user satisfaction \cite{tatsunori2018repeated} or AUC \cite{preethi200adversarially}). This branch of methods may not be very competitive in fairness performance with evaluations such as Equalised Odds and Demographic Parity, as these metrics are designed for Group Fairness methods. However, as mentioned before, some datasets may contain biased labels, and focusing on increasing classification performance for the minority may lead to an increase in the bias gap, which is what we aim to avoid. Therefore, our study underscores the significance of group fairness for a more comprehensive improvement, steering away from inadvertently perpetuating bias in labelled datasets.

\subsection{Ablation Study}\label{sec:ablation}

In our ablation study, we look at the effectiveness of the two components in the proposed framework, Reckoner, and the necessity of combining them. Tables~\ref{table1} - ~\ref{table3} show prediction accuracy and fairness measurements for all variants. In general, Reckoner and its variants  achieve superior performance on the COMPAS and NewAdult datasets. In the CelebA dataset, Reckoner and its variants  exhibit better performance only in Demographic Parity compared all the baselines. Based on these results, we are interested in understanding the differences in performance among each variant across different datasets.

\textbf{Effect of the Learnable Noise.} Our model without learnable noise trains both classifiers using original inputs in the refinement stage. It achieves a slight advantage in accuracy on COMPAS and CelebA datasets, with a minor gap in fairness results. However, in the New Adult dataset there is no particularly noticeable difference in performance across the three metrics compared with the full Reckoner. It is worth noting that we argue that COMPAS and CelebA datasets contain biased labels, as their ground truth is assigned by officials of the legal department and human annotators. 
Due to the absence of learnable noise, the classifier in the \textit{pseudo-learning} phase tends to push the decision boundary significantly into the feature space of the majority, enhancing discrimination against the minority and leading to an increase in accuracy. However, when the dataset labels are considered biased, this can also result in a decrease in fairness values. This is why in the New Adult dataset, whether the proposed framework introduces learnable noise or not, the performance difference is not significant. However, in datasets suspected of having biased labels there is a larger fluctuation in fairness among these two variants of the proposed approach.

\textbf{Effect of the \textit{pseudo-learning}.} In addition to the learnable noise, we also evaluate the effectiveness of \textit{pseudo-learning}. Our model without \textit{pseudo-learning}  employs a three-layer MLP as the main classifier, takes noise-augmented information as input, and does not involve the identification stage. Similar to our previous speculation, its poor performance on datasets with biased labels is also related to the misguidance caused by supervised learning with problematic ground truth. Without the regularisation provided by \textit{pseudo-learning}, the classifier mistakenly treats bias as knowledge, and learnable noise reinforces this by selecting features that maximise the cross-entropy with the ground truth. Additionally, experimental results seem to suggest overfitting issues with this variant on these datasets. As compared to the results on New Adult, where our method outperforms five baselines and other variants, both in terms of predictive accuracy and fairness, these results suggest that the bias in labels will greatly impact algorithmic fairness. Perhaps in a dataset with unbiased labels, improving predictive accuracy could effectively reduce the bias gap between different demographic groups.

\section{Related Work}

\textbf{Group Fairness.} In contrast to approaches that emphasise the equitable treatment of similar individuals in pursuit of individual fairness, our work focuses on group fairness, manifesting in the differential treatment of distinct demographic groups. Some foundational methods to group fairness include incorporating fairness regularisation into the objective function or converting it into a constrained optimisation problem. \citet{kamishima2011fairness} introduced a method for reducing mutual information between sensitive groups and targets by quantifying the mutual distribution between them. This approach aims to diminish the dependency between sensitive groups and targets. A similar concept is also adopted by \citet{beutel2019putting}, where fairness is achieved by minimising the absolute correlation between these two entities. In contrast to the aforementioned methods, \citet{hardt2016equality} propose the use of the Equalised Odds fairness metric, which underscores the equalisation of true positive and false positive rates across different demographic groups. It transforms the general loss function into an optimisation problem subject to fairness constraints, ensuring that the revised unbiased predictions closely approximate the original predictions. Similarly, \citet{zafar2019fairness} achieve fair classification by adding tractable constraints at the decision boundary. However, as the desire for both algorithmic fairness and privacy grows, we observe the requirement of avoiding the use of sensitive attributes in machine learning model training, leading to legislative restrictions on such practices like, e.g., the General Data Protection Regulation (GDPR)~\cite{voigt2017eu}. To manage such requirements, some approaches have been designed under the assumption that sensitive attributes are either difficult to obtain or prohibited from use, like we do in our work.

\textbf{Fairness Without Sensitive Attributes.} As public concerns about privacy are on the rise, an increasing amount of research on group fairness is turning its attention to ``imperfect" data, such as missing protected class labels \cite{chen2019fairness, awasthi2021evaluating, liu2023group} or noisy sensitive attributes \cite{ghazimatin2022measuring, mehrotra2021mitigating}. To deal with fairness  in this setting, the main idea of some studies is leveraging the correlation between sensitive and non-sensitive attributes to mitigate bias. Representative work includes the use of proxy features \cite{gupta2018proxy}, in which a proxy group is obtained from clustering the data and is used to replace actual sensitive attributes during training. A well-known example is using `zip code' instead of `race' as this can have similar effects on individual splits since the two attributes are highly correlated \cite{datta2017proxy}. Similarly, \citet{tianxiang2022related,shen2020class} explore features which have strong correlation with sensitive attributes to learn fair classifiers by using them for training and for regularisation in learning. However, this approach needs a careful selection of proxy attributes and even of fairness metrics. To address underlying issues, \citet{zhaowei2023weak} estimate fairness using only weak proxies. Through estimating the transition probabilities between sensitive group target values, it uses auxiliary models to calibrate the fairness metrics. Another family of approaches \cite{tatsunori2018repeated} addresses fairness without sensitive attributes via distributionally robust optimisation (DRO). The main idea is that the fairness of the algorithm is related to the quantity of individuals in different demographic groups. If empirical risk minimisation is employed to optimise the algorithm, it may lead to a prioritisation of benefits for the privileged group, as the privileged group constitutes the majority. This could result in the non-privileged group gradually avoiding the algorithm due to a poor experience, creating a vicious cycle that ultimately causes the algorithm to increasingly overlook the non-privileged group. This approach utilises different group proportions to design upper and lower bounds for the objective function of each group, ensuring that no group is discriminated against by the algorithm. Recently, \citet{DBLP:conf/iclr/JungPCM23}, targeting group fairness, extended DRO with fairness constraints in the resulting objective function using a re-weighting based learning method. 
Beside the aforementioned methods, others have recently utilised various techniques to address unfairness without  knowledge of demographics. For example, \citet{preethi200adversarially} adversarially reweigh the samples to achieve a Rawlsian Max-Min fairness and learn the classifier. However, these methods can be easily influenced by outliers. Others tackle the problem through knowledge distillation \cite{junyi2022knowledge}, reweighing-based contrastive learning \cite{chai2022self}, and causal variational autoencoders \cite{vincent2022causal}. However, these methods need the prior identification of proxies to harness their interactions with sensitive attributes, such as correlation and causality, in order to achieve fairness. Our approach avoids the need for such analysis. Instead, it leverages learnable noise applied to all data, forcing the data to retain only essential information for better predictions. Additionally, it employs a dual-model knowledge-sharing mechanism to acquire fairness-related knowledge, thereby improving predictive fairness. Hence, our proposed framework exhibits greater generalisability, particularly when dealing with data where proxy identification is challenging, such as images and audio.

\section{Conclusions}
In this paper, we present a novel framework for classification tasks that improves fairness without using sensitive attributes. Through an analysis of the bias gap and the distribution of selected non-sensitive attributes across different confidence subsets with respect to different demographic groups, we gain insights into how biased labels and other hidden bias harm fairness in predictions and mislead the classifier. Our proposed framework integrates (1) learnable noise and (2) a dual-model system, enabling a knowledge-sharing framework for fair predictions for different demographic groups. Our experimental results show the superiority of the proposed method, which can make accurate and fair predictions, as compared to state of the art baseline methods. Our ablation study also confirms the benefits of the two main components in our proposed solution. The code of our method is publicly available at \url{https://github.com/uewopq88/Reckoner-Fairness}. It is important to acknowledge that the critical issue of the intersectionality of multiple sensitive attributes in automated decision-making systems warrants further investigation in our future research. While our proposed framework is able to perform classification tasks even with multiple sensitive attributes, this paper does not discuss intersectional fairness due to challenges such as data scarcity at intersections of minority groups and a lack of proper group fairness metrics. Future studies will focus on enhancing intersectional fairness in classification and extending current group fairness metrics.

\begin{acks}
This work is partially supported by the Australian Research Council (ARC) Training Centre for Information Resilience (Grant No. IC200100022).
\end{acks}

\bibliographystyle{ACM-Reference-Format}
\bibliography{sample-base}

\end{document}